\documentclass[twocolumn]{article}
\usepackage[letterpaper,top=2cm,bottom=2cm,left=2cm,right=2cm,marginparwidth=1.75cm]{geometry}
\usepackage{microtype}
\usepackage{graphicx}
\usepackage{graphics}
\usepackage{booktabs} 
\usepackage{caption}
\usepackage{subcaption}
\usepackage{hyperref}
\hypersetup{colorlinks,citecolor=blue, linkcolor=blue,urlcolor=blue}
\usepackage{amsmath}
\usepackage{amssymb}
\usepackage{mathtools}
\usepackage{amsthm}
\usepackage{float} 
\usepackage{bbm}
\usepackage{natbib}
\usepackage[parfill]{parskip}
\theoremstyle{plain}

\theoremstyle{definition}

\theoremstyle{remark}

\usepackage{times}
\usepackage{changepage}
\usepackage{authblk}
\usepackage[textsize=tiny]{todonotes}

\title{
    \vspace{-1cm}
    \linethickness{1pt}\line(1,0){500}\\
    \vspace{0.5cm}
    \textbf{Deep de Finetti: Recovering Topic Distributions \\from Large Language Models}\\
    \linethickness{1pt}\line(1,0){500}
}

\author{\textbf{Liyi Zhang \quad R. Thomas McCoy \quad Theodore R. Sumers \quad Jian-Qiao Zhu \quad Thomas L. Griffiths}}
\date{\textit{Princeton University \\
    \{zhang.liyi, tom.mccoy, sumers, jz5204, tomg\}@princeton.edu}}

\begin{document}

\maketitle

\begin{abstract}
\begin{adjustwidth}{0.6cm}{0.6cm}
Large language models (LLMs) can produce long, coherent passages of text, suggesting that LLMs, although trained on next-word prediction, must represent the latent structure that characterizes a document. Prior work has found that internal representations of LLMs encode one aspect of latent structure, namely syntax; here we investigate a complementary aspect, namely the document’s topic structure. We motivate the hypothesis that LLMs capture topic structure by connecting LLM optimization to implicit Bayesian inference. De Finetti’s theorem shows that exchangeable probability distributions can be represented as a mixture with respect to a latent generating distribution. Although text is not exchangeable at the level of syntax, exchangeability is a reasonable starting assumption for topic structure. We thus hypothesize that predicting the next token in text will lead LLMs to recover latent topic distributions. We examine this hypothesis using Latent Dirichlet Allocation (LDA), an exchangeable probabilistic topic model, as a target, and we show that the representations formed by LLMs encode both the topics used to generate synthetic data and those used to explain natural corpus data.
\end{adjustwidth}
\end{abstract}


\section{INTRODUCTION}
\label{sec:introduction}

In order to understand large language models (LLMs), it is important to determine what information they encode in their internal representations. One way to approach this question is by considering what LLMs model: text documents. Documents can be thought of as sequences of words, but beneath the surface they also have several types of latent structure, such as the syntax that shapes sentences and the topics that underlie the information in the text. LLMs excel at producing high-quality documents, suggesting that they must somehow capture this latent structure. Indeed, a rich body of literature (reviewed in Section~\ref{sec:probing}) has found evidence that the hidden states of LLMs encode many latent variables that underlie language, including syntax \citep{hewitt2019structural}, world states \citep{li2021implicit}, and agent stances \citep{andreas-2022-language}. 



In this work, we hypothesize that LLMs encode a document’s topic structure---a latent variable complementary to those studied in prior work (although suggested by \citet{andreas-2022-language}).
We motivate this hypothesis through an analysis of the LLM training task of autoregression (predicting the next word in a passage of text).
An extension of de Finetti's theorem shows that, for sequences that are exchangeable, 
learning to fit the autoregressive distribution is implicitly equivalent to conducting Bayesian inference on the latent distribution underlying the data \citep{korshunova2018bruno}, 
\begin{align}
p(x_{n+1}|x_{1:n}) & = \int_\theta p(x_{n+1}|\theta)p(\theta|x_{1:n}) d\theta,
\label{eq:gpt_to_bayes}
\end{align}
where $\theta$ denotes latent generating distributions and the sequence $x_1, x_2, ..., x_n$ is said to be exchangeable if changing the order of the words in $x_{1:n}$ does not affect $p(x_{n+1}|x_{1:n})$.
Linguistic sequences are not, in fact, exchangeable. However, certain latent properties of text---such as the topic structure---depend little on word order. Therefore, we propose that Equation \ref{eq:gpt_to_bayes} can be used to understand why LLMs trained on next-word-prediction implicitly learn a document's topic structure.

Our experiments support the hypothesis that LLMs execute implicit Bayesian inference: the latent distribution $p(\theta | x_{1:n})$ for certain $\theta$ are recoverable from LLMs using a linear probe.
We first conduct experiments in a synthetic setting where we train LLMs on artificial exchangeable sequences, thereby fully satisfying the assumptions underlying de Finetti's theorem. In this setting, as predicted by de Finetti's theorem, the latent topic distribution $p(\theta | x_{1:n})$ can be decoded from LLMs with high accuracy. 
We then move to models trained on natural text, where de Finetti's theorem no longer strictly applies. While we no longer have theoretical guarantees that LLMs should learn the latent topic distribution, we hypothesize that the topic distribution will still be decodable because topic structure depends little on word order --- even though text in its full complexity is not exchangeable, at the level of topics it may still be reasonably approximated as such.  
Taking Latent Dirichlet Allocation (LDA; \citet{lda}) as a proxy Bayesian model of topic structure, we show that the encodings of three LLMs (\textsc{Gpt-2}, \textsc{Llama 2}, and \textsc{Bert}) contain information analogous to that extracted by LDA, supporting the hypothesis that LLMs implicitly perform Bayesian inference even in a natural setting where the assumptions behind de Finetti's theorem do not strictly hold. 

Our results thus provide two contributions: First, we show that LLMs trained on next-word prediction also learn documents' topic mixtures. Second, we explain this success by linking it to de Finetti's theorem, which suggests that autoregressive LLMs implicitly represent latent generating distributions. 


\section{BACKGROUND}
\label{sec:background}

\subsection{Analyzing Large Language Models}\label{sec:probing}

An extensive literature has investigated the information encoded in the vector representations of language models \citep{gupta2015distributional,kohn2015whats,ettinger2016probing,adi2017finegrained,hupkes2018visualisation}; for overviews, see \citet{rogers2020primer} and \citet{belinkov2022probing}. Much of this literature focuses on testing whether language models encode syntactic information such as part of speech \citep{shi2016string,belinkov2017neural}, syntactic number \citep{giulianelli2018hood,conneau2018cram}, or sentence structure \citep{tenney2018what,hewitt2019structural,liu2019linguistic,lin2019open}. Several works have decoded semantic properties such as entity attributes \citep{gupta2015distributional,grand2022semantic}, sentiment \citep{radford2017learning}, semantic roles \citep{ettinger2016probing,tenney2018what}, world states \citep{li2021implicit}, and agent properties \citep{andreas-2022-language}. A few works have decoded the semantic property that we study, namely, topic structure \citep{Li2023HowDT,Sia2020TiredOT,Meng2022TopicDV,Zhang2022IsNT}. Of these, \citet{Sia2020TiredOT, Meng2022TopicDV, Zhang2022IsNT} use \textsc{Bert} for the sake of topic discovery. However, we equate the language modeling objective with Bayesian inference and show that LLMs mimic LDA. \citet{Li2023HowDT} also performs model-based probing on LLMs to find encodings of topics. However, \citet{Li2023HowDT} compares pairwise embeddings of words that belong to the same topics and those that belong to different topics. Our work complements their work by probing the topic mixture vector, which connects more directly to the idea of equating language modeling with Bayesian inference.

Like our work, three prior papers have also analyzed LLMs as implicitly learning to perform Bayesian inference---namely, \cite{xie2021explanation}, \cite{mccoy2023embers}, and \cite{wang2023large}. Of these, the most closely related to ours is \cite{wang2023large}, which also connected LLMs to topic models. There are two important differences between these works and ours. First, the focus of our experiments is analysis of internal representations, which none of these prior papers performed; they instead analyzed LLMs at the behavioral level, making our work complementary to theirs. 
Second, the goal of these prior papers was to characterize when and/or why LLMs succeed at performing in-context learning of new tasks. 
Our focus is instead on analyzing how LLMs perform what they were trained to do (modeling documents), rather than analyzing how this ability might be co-opted to learn new tasks.
For instance, in \citeauthor{wang2023large}'s analysis of LLMs as topic models, the relevant notion of ``topic'' is a task to be learned in-context, whereas the topics that we study are instead of the type used in traditional topic models (i.e., the topic that a document is about). 

\subsection{Topic Modeling}

\begin{figure}[h]
    \centering
    \includegraphics[width=0.4\textwidth]{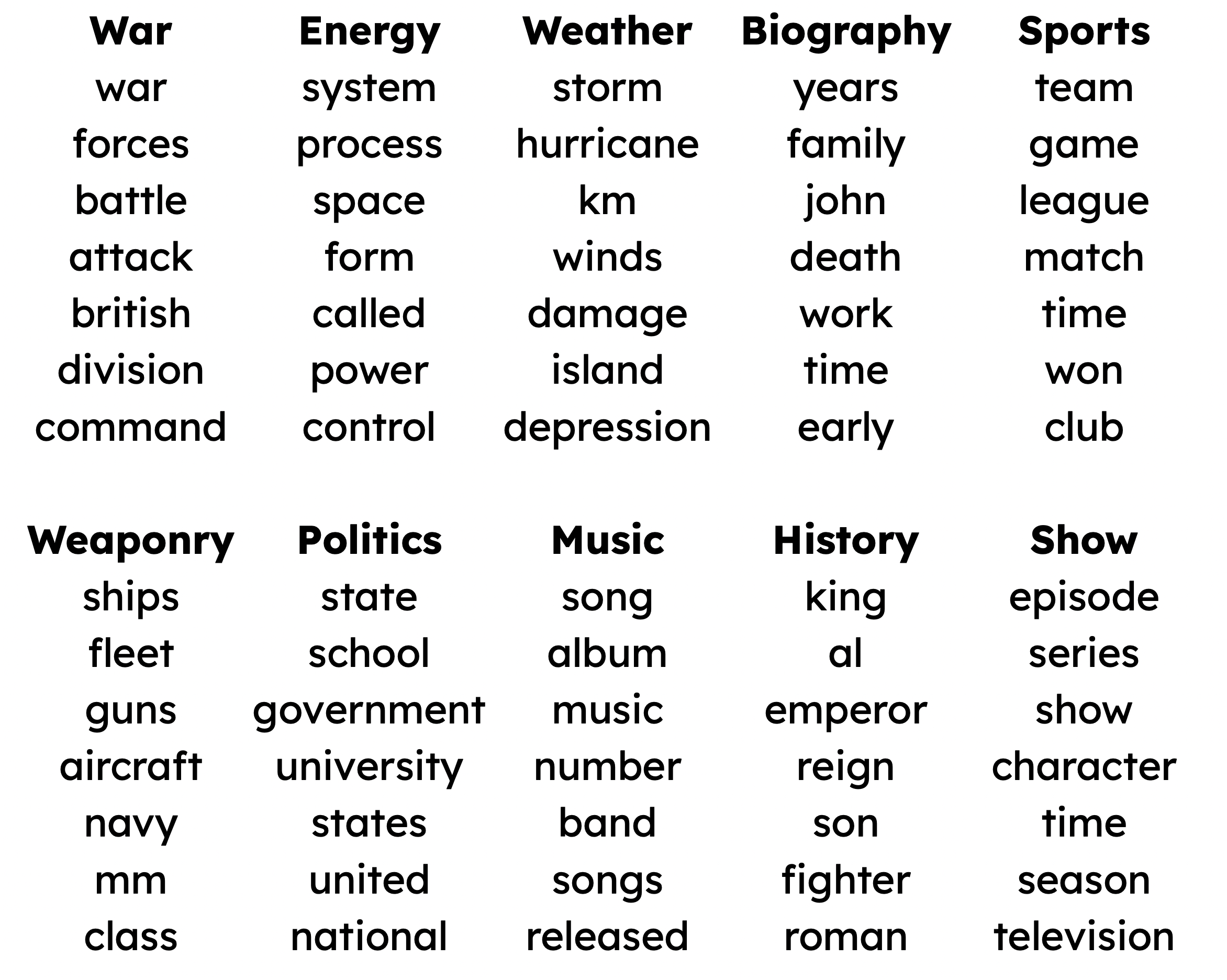}
    \caption{Inferred topics from an LDA model used in empirical evaluations. These are top seven words from 10 out of the 20 topics derived from the WikiText-103 dataset after weighting via TFIDF. We manually summarized each topic with a word, written in bold. We define a document's topic structure as a mixture of topics such as these; we use such topic mixtures as targets for decoding from LLM representations.}
    \label{fig:lda-topic}
\end{figure}

Because our goal is to model topics, we need a way to formally characterize a topic. For this purpose, we use Latent Dirichlet Allocation (LDA; \citet{lda}), a generative model that is widely used for modeling the topic structure of documents. A document is assumed to be generated from a mixture of $K$ topics. Each topic is a distribution over the vocabulary; e.g., a topic corresponding to geology might assign high probability to words such as \textit{mineral} or \textit{sedimentary}. The resulting generative model is,
\begin{enumerate}
    \item For each topic $k$ in $(1,...,K),$
    \begin{enumerate}
        \item Draw topic $\beta_k \sim \text{Dirichlet}_V(\eta)$.
    \end{enumerate}
    \item For each document $i$,
        \begin{enumerate}
            \item Draw topic mixture $\theta_i \sim \text{Dirichlet}(\alpha).$
            \item For each word $j$ in document $i$,
            \begin{enumerate}
                \item Draw topic assignment \\$t_{ij} \sim \text{Categorical}(\theta_i)$,
                \item Draw word $x_{ij} \sim \text{Categorical}(\beta_{t_{ij}})$,
            \end{enumerate}
        \end{enumerate}
\end{enumerate}

where $V$ is the vocabulary size, and $\alpha$ and $\eta$ are pre-initialized hyperparameters. After LDA is trained on a corpus, the inferred quantities can be used to explore the corpus. The latent variable $\theta_i$ stands for each document's underlying topic mixture. An inferred latent variable commonly used to visualize and understand a topic model is the vector $\beta_k$, which is a distribution over the vocabulary in topic $k$. As an example, Figure \ref{fig:lda-topic} shows the top seven entries from $\beta_k$, $k=1,2,...,10$, i.e., the top seven words from ten inferred topics, weighted by a TFIDF term score that de-weights words that are common in every topic \citep{lda}.

Our core hypothesis in this work is that LLMs implicitly encode the topic structure of a document. The LDA framework gives us a way to make this hypothesis precise: we operationalize encoding a document's topic structure as encoding the $\theta_i$ vector that underlies document $i$ in an LDA model. This framing provides a concrete way to test this hypothesis: investigating whether $\theta_i$ vectors can be decoded from LLM document representations.

\subsection{De Finetti's Theorem}
\label{sec:2.3}

De Finetti's theorem shows that any exchangeable distribution can be expressed in terms of independent sampling conditioned on a latent generating process. 
More formally, given an exchangeable process $x_1, x_2, ..., x_n$, we can model the joint distribution as a mixture of i.i.d. samples,
\begin{align}
    p(x_1, ..., x_n) = \int_{\theta} p(\theta)\prod_{i=1}^np(x_i|\theta) d\theta
\label{eq:definetti}
\end{align}
where $\theta$ are the parameters of the generating process $p(x_i|\theta)$ and a sequence $x_1, x_2, ..., x_n$ is said to be exchangeable if $p(x_1, ..., x_n) = p(x_{\pi(1)}, ..., x_{\pi(n)})$, for any $n>1$ and any permutation $\pi(n)$.
In other words, the joint distribution remains invariant under any permutation of the sequence.
All sequences that are i.i.d. are also exchangeable; however, the converse does not hold, and exchangeable random variables may exhibit correlation. 

To further elucidate the relationship between an autoregressive distribution and Bayesian inference, we use an extension of Equation \ref{eq:definetti} \citep{korshunova2018bruno},
\begin{align*}
p(x_{n+1}|x_{1:n}) & = \int_\theta p(x_{n+1}|\theta)p(\theta|x_{1:n}) d\theta.
\end{align*}
Here, de Finetti's theorem connects learning to fit the autoregressive distribution (the left hand side of Equation \ref{eq:gpt_to_bayes}) with conducting Bayesian inference on the latent generative process underlying the data (the right hand side of Equation \ref{eq:gpt_to_bayes}). That is, the autoregressive distribution can be seen as the posterior predictive distribution in Bayesian modeling.



\section{APPROACH}
\label{sec:method}

\subsection{LLMs and Exchangeability}

\subsubsection{Autoregressive Language Models}

To explain LLMs' capability to capture the topic structure of a document, we connect the \textbf{autoregressive LLM objective} with implicit Bayesian modeling using de Finetti's theorem. First, we write out the autoregressive objective,
\begin{align}
    \mathcal{L}_{ALM}(x_{1:N}) &= \log p(x_{1:N})\nonumber \\
    &= \log p(x_1) + \sum_{n=1}^{N-1} \log p(x_{n+1}|x_{1:n}),
\end{align}
where $x_{1:N}$ is the sequence of words of the document. Using Equation \ref{eq:gpt_to_bayes}, which is the equation that sets the equivalence between exchangeable models and implicit Bayesian inference, we have
\begin{align}
    \mathcal{L}_{ALM}(x_{1:N}) = &\log p(x_1) \nonumber \\
    &+ \sum_{n=1}^{N-1} \log \int_\theta p(x_{n+1}|\theta)p(\theta|x_{1:n})d\theta.
\end{align}
In other words, modeling a sequence of words with an exchangeable model is equivalent to performing Bayesian inference on each word. 

Language sequences, however, are not exchangeable; the interdependence of words within a sentence is crucial for semantic interpretation. Recognizing this complexity, we explore the nuanced ways in which certain facets of language might be treated differently. It may be plausible to regard specific aspects or levels of language data as partially exchangeable \citep{diaconis1980finetti}. In partial exchangeability, subsets of random variables are exchangeable within themselves, but not necessarily across different subsets. In localized contexts, word order may not matter. For example, the sequences ``apple, banana, cherry'' and ``banana, cherry, apple'' could be semantically equivalent when enumerating a list of fruits. Within such contexts, the words qualify as partially exchangeable. At a more abstract level, text segments that represent discrete semantic units (e.g., sentences within a paragraph) might also be considered partially exchangeable if they contribute independently to the overall meaning, contingent upon the topic. 


\subsubsection{Masked Language Models}

Masked language models (MLMs) are another class of LLMs that have been successful at modeling text, and we explore whether the objective of MLM is also equivalent to implicit Bayesian inference under the exchangeability assumption. We find that it is, but it differs from the breakdown for autoregressive language models and results in less expressivity.

MLMs, notably including the \textsc{Bert} series \citep{devlin-etal-2019-bert}, randomly mask certain tokens in the input sequence, and the model's goal is to predict the original words at the masked indices only by considering their surrounding context. Unlike autoregressive models, the MLM does not model a coherent joint distribution of the data \citep{yamakoshi-etal-2022-probing, young2022inconsistencies}. However, the log objective can be extended as follows,
\begin{align}
    \mathcal{L}_{MLM}(x_{1:N}) &= \sum_{n \in M} \log p(x_n | x_{i,i\in U}) \nonumber \\
    &= \sum_{n \in M} \log \int p(x_n | \theta)p(\theta|x_{i,i\in U})d\theta. \label{eq:mlm-bayes}
\end{align}
where $M$ denotes the set of masked indices, and $U$ denotes the set of unmasked indices. The proof is given in Appendix \ref{sec:appendix_proof}, and is a simple extension of the derivation for the autoregressive version. The difference between the MLM objective and the autoregressive objective is that in the summation, the prediction of each token $x_n$ uses the same posterior over the latent variable $p(\theta|x_{i,i\in U})$. In other words, each token $x_n$ is predicted independently from the latent variable $\theta$. As a result, MLM forms a less expressive Bayesian inference objective than autoregressive models. Therefore, we aim to empirically evaluate both the ability of MLM to recover latent variables, and whether its performance forms a contrast against that of autoregressive models.


\subsection{Probing for Topic Mixtures}
\label{sec:3.2}

In this section we discuss the methodology for recovering topic mixtures from LLMs. 

LLMs conduct downstream tasks, including predicting the next word $x_{N+1}$, via an embedding of the seen sequence $f(x_{1:N})$. Based on the analysis above, we expect $f(x_{1:N})$ to be sufficient to reconstruct the distribution over topics given observed data $p(\theta|x_{1:n})$. Concretely, for each document $i$, we define the input as the LLM-learned document embedding $f(x_{1:N})_i$. We define the target as a topic mixture $\theta_i$ that is sampled from $p(\theta|x_{1:n})$ learned by LDA. Thus, LDA is a proxy for ground truth topic distribution. This target is analogous to a soft-label classification target. The topic probe maps from the document embedding to the topic mixture target. To ensure that statistical model learning is contained in the LLM, not in the probe, we keep the probe simple by defining it as a linear classifier with softmax activations. In summary, given a natural corpus and a pretrained LLM, the steps to probe topics are,
\begin{enumerate}
    \item Train LDA with $K$ topics on the corpus.
    \item For each document $i$,
    \begin{enumerate}
        \item Get LLM embedding $f(x_{1:N})_i$.
        \item Get LDA topic mixture $\theta_i$.
    \end{enumerate}
    \item Train topic probe $g$ by minimizing the cross-entropy loss: $-\frac{1}{N}\sum_{i=1}^N \sum_{k=1}^K \theta_{i_k} \log g(f(x_{1:N})_i)_k $.
\end{enumerate}
On synthetic data, however, the LLM is instead trained on data sampled from an LDA model that is manually defined (i.e., not trained), and the topic probe targets the ground-truth topic mixtures instead of topic mixtures that are estimated by a trained LDA model. The steps for synthetic data generation and exploration are detailed in Section \ref{sec:4.1}.

\textbf{Implementation.} The document embedding $f(x_{1:N})$ is derived from the LLM embedding for each token of the sequence, which is by default the representation of the final layer before the LLM prediction head. In an experiment, we also search across layers as the document embedding. The options for document embedding $f(x_{1:N})$ are the embedding of each individual token in the final layer, as well as an average of all token embeddings in this layer. In each case study, we choose the one that gives best topic recovery performance. In general, for both autoregressive models and MLM, either the last token embedding or the average gives best performance, depending on the dataset.

\begin{figure*}
    \centering
    \includegraphics[width=15cm]{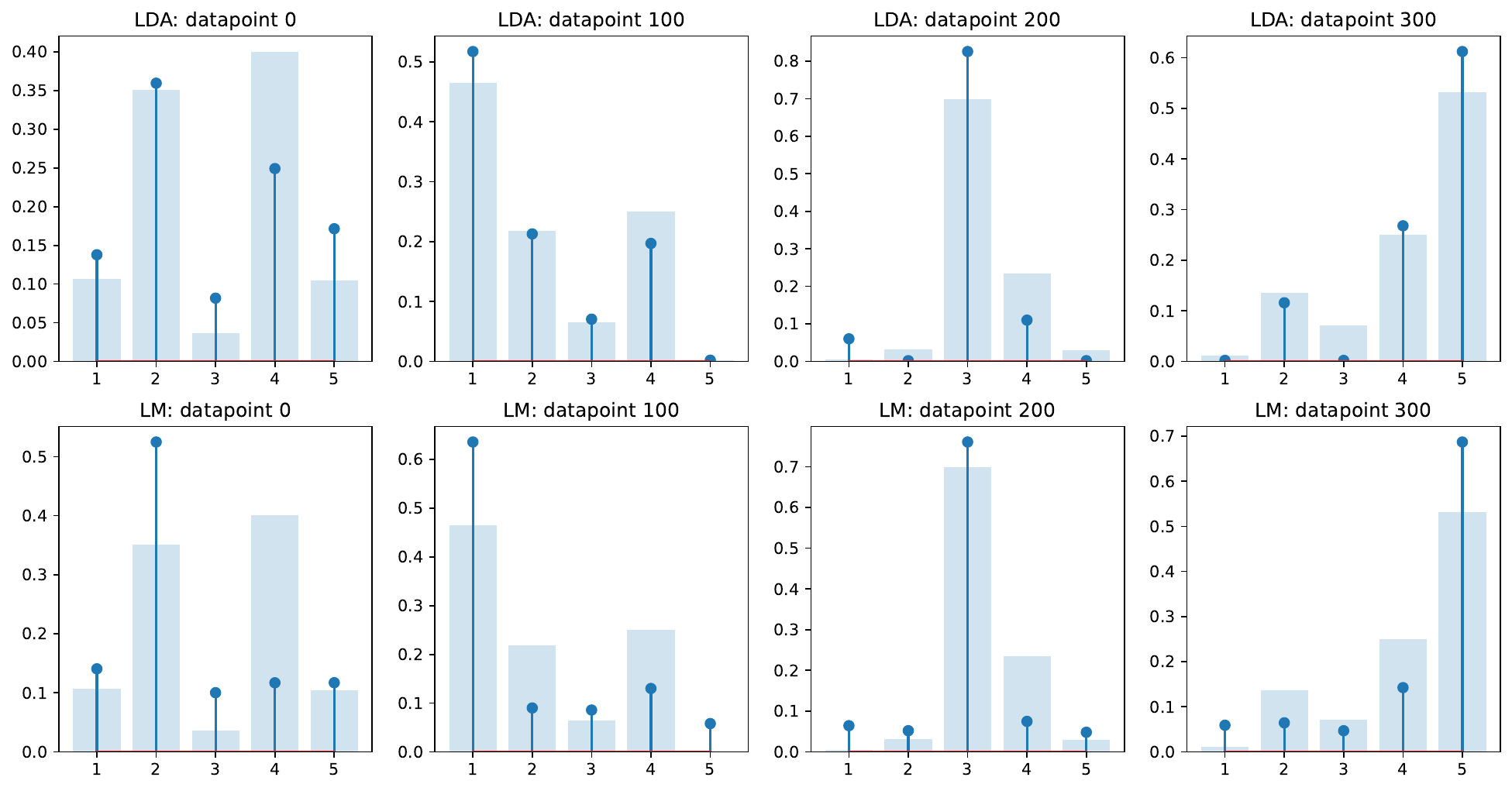}
    \caption{Distribution of synthetic data topics predicted by LDA (row 1) and AT classifier (row 2) for different validation datapoints. Predictions are stick plots, and ground truth is bar chart in the background. They both exhibit learning of topic mixtures by trying to match the distribution, in addition to top-1 agreement.}
    \label{fig:dist}
\end{figure*}


\section{EXPERIMENTS}
\label{sec:experiments}
We empirically evaluate the extent to which LLMs recover topic mixture on three datasets: a synthetic dataset, \textit{20Newsgroups} (20NG), and \textit{WikiText-103}. 

\begin{table}[h]
\centering
\caption{Topic prediction performance of the autoregressive transformer (AT), \textsc{Bert}, LDA, and word-embedder (WE) on the synthetic datasets. Hyperparameter $\alpha$ defines the dataset generation process, where a higher $\alpha$ means a more difficult task with underlying topics being more evenly distributed. AT and \textsc{Bert} have similar performance in the easiest setting, but AT performs well in harder settings where \textsc{Bert} performances worsen. End-to-end WE achieves stronger performance than language models, and LDA matches expectations by providing an upper bound in performance.}
\resizebox{1\columnwidth}{!}
{\begin{tabular}{ccccc} 
 \toprule
 $\alpha$ & Method & Accuracy $\uparrow$ & L2 loss $\downarrow$ & Tot. var. loss $\downarrow$\\
 \midrule
 & AT & $82.8\% \pm 0.5\%$ & $0.041 \pm 0.001$ & $0.141 \pm 0.001$ \\
 0.5 & \textsc{Bert} & $83.6\% \pm 1\%$ & $0.036 \pm 0.003$ & $0.131 \pm 0.005$\\
 & LDA & $87\% \pm 0.6\%$ & $0.029 \pm 0$ & $0.117 \pm 0.001$\\
 & WE & $85.8\% \pm 1.3\%$ & $0.03 \pm 0.001$ & $0.119 \pm 0.002$ \\
 \midrule
 & AT & $75.5\% \pm 0.8\%$ & $0.044 \pm 0.001$ & $0.144 \pm 0.001$ \\
 0.8 & \textsc{Bert} & $51.5\% \pm 1.7\%$ & $0.111 \pm 0.005$ & $0.233 \pm 0.011$\\
 & LDA & $82.6\% \pm 0.5\%$ & $0.036 \pm 0.001$ & $0.133 \pm 0.004$\\
 & WE & $80.9\% \pm 0.5\%$ & $0.029 \pm 0$ & $0.116 \pm 0.001$ \\
 \midrule
 & AT & $70.5\% \pm 1.6\%$ & $0.045 \pm 0.001$ & $0.146 \pm 0.003$ \\
 1 & \textsc{Bert} & $46.6\% \pm 3.3\%$ & $0.1 \pm 0.004$ & $0.222 \pm 0.006$\\
 & LDA & $79.6\% \pm 1.4\%$ & $0.045 \pm 0.004$ & $0.147 \pm 0.006$\\
 & WE & $79.4\% \pm 1\%$ & $0.027 \pm 0$ & $0.113 \pm 0.001$ \\
 \bottomrule
\end{tabular}}
\label{tab:toy-lm-lda}
\end{table}

\begin{figure}[h!]
    \centering
    \begin{subfigure}[b]{0.215\textwidth}
        \centering
        \includegraphics[width=\textwidth]{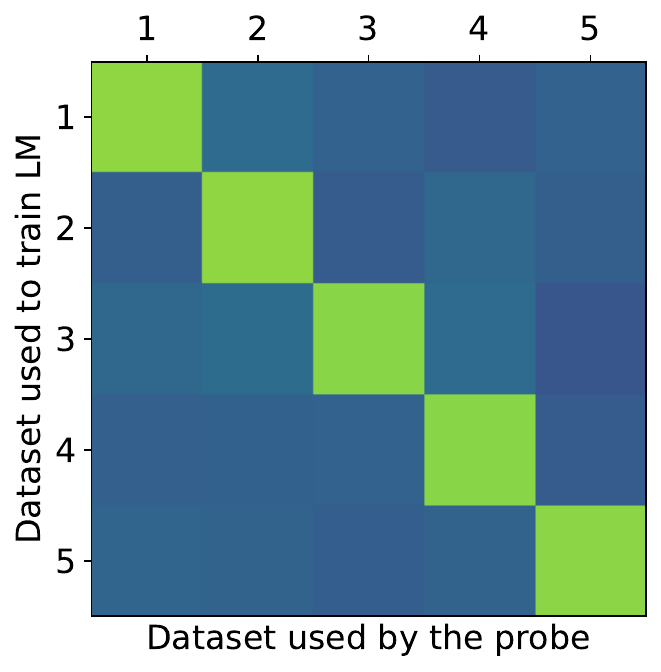}
        \caption{AT probe accuracies.}
        \label{fig:heatmap-lm}
    \end{subfigure}
    \begin{subfigure}[b]{0.2577\textwidth}
        \centering
        \includegraphics[width=\textwidth]{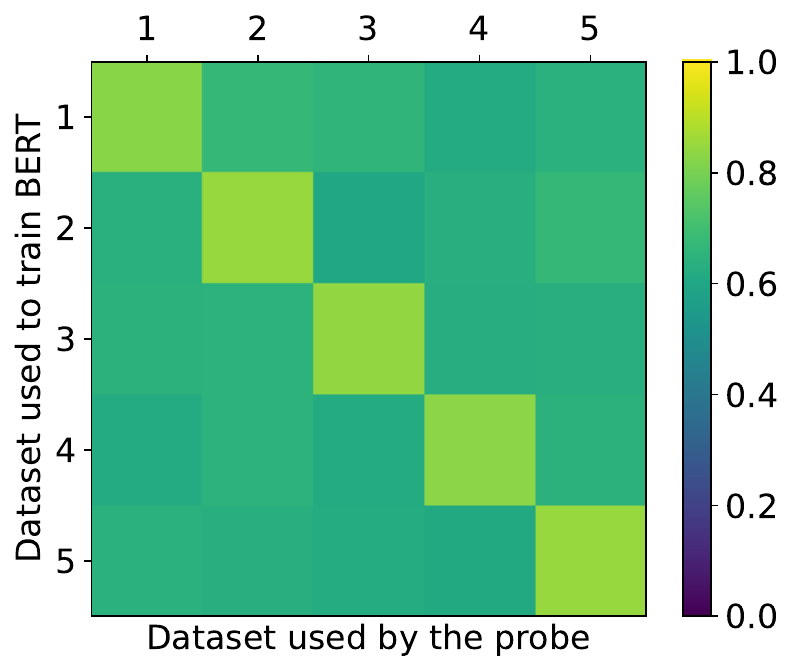}
        \caption{\textsc{Bert} probe accuracies.}
        \label{fig:heatmap-bert}
    \end{subfigure}
    \hfill
    \caption{Control experiments showing AT's (\ref{fig:heatmap-lm}) and \textsc{Bert}'s (\ref{fig:heatmap-bert}) topic probe validation performance on synthetic data. For each of AT and \textsc{Bert}, five models are trained and validated on five datasets, with each dataset generated by a distinct topic model. Colors show probe accuracy. A cell on row $i$ and column $j$ corresponds to model $i$ on dataset $j$, so the diagonal corresponds to a model on its own dataset. {\it For AT, performance is only strong on the dataset with the same generating topic model, suggesting that the underlying statistical model, not the probe taking different word embeddings, is responsible for performance -- a relationship that is weaker for \textsc{Bert}.}}
    \label{fig:heatmap-synthetic}
\end{figure}

\subsection{Synthetic Data}
\label{sec:4.1}

\textbf{Experiment description.} A synthetic dataset is bags-of-words generated by a manually initialized topic model. We set the vocabulary size $V=10^3$, number of topics $K=5$, and generated $N=10^4$ documents that are each 100 words long. The steps made in this case study are,
\begin{enumerate}
    \item Draw bags of words \textbf{x} from topic model, $\textbf{x} \sim \text{LDA}(\alpha, \eta)$.
    \item Train a language model on corpus \textbf{x}.
    \item Train a linear classifier that takes the language model embedding for each document, and targets ground truth topic proportions $\theta$. (The language model weights are fixed in this step.)
\end{enumerate}
Unlike the experiments on natural corpora, the language models were initialized at random and trained on the dataset. Additionally, the topic probe was trained to target the ground truth topic mixtures rather than those inferred by LDA.

\subsubsection{Topic Prediction} 

\textbf{Models.} We trained four models: an autoregressive transformer decoder (AT), \textsc{Bert}, LDA, and an end-to-end word embedder (WE). Since an LLM with hundreds of millions of parameters is too large for this dataset, we implemented a transformer decoder that performs autoregressive prediction similar in style to, for instance, \textsc{Gpt-2}. 
For MLM, we implemented a small version of \textsc{Bert} called \textsc{Bert-tiny} \citep{Turc2019WellReadSL}, which has 608,747 parameters when vocabulary size $V=10^3$. This model size corresponds to AT with four decoder layers. 
LDA is implemented to establish an upper-bound for model performance. 
We also included a model intended to provide an upper bound for embedding performance: a word embedder which is a matrix that maps from the vocabulary space to the AT / \textsc{Bert} embedding dimension and is end-to-end trained with the probe that predicts topics. 

\textbf{Metrics.} To measure the recovery of topic mixtures, where in addition to correct classification, an accurate prediction of spread is required, we use accuracy, cross-entropy loss, L2 loss, and total variation loss. The metrics are used on the predicted topic mixture and the ground truth topic mixture that generated the document. Accuracy is defined as how often the top topic predicted by the classifier's mixture agrees with the top topic from ground truth. The remaining loss measures apply to the whole topic vector.

\textbf{Implementation.} Hyperparameters are detailed in Appendix \ref{sec:appendix_implementation}. All training procedures used the Adam optimizer \citep{Kingma2015AdamAM} and targeted the cross-entropy loss. For each model, learning rate is tuned in $[0.00003, 0.001]$, and batch-size in $[8, 128]$. For each probe, learning rate is tuned in $[0.0001, 0.03]$, and batch-size in $[8, 64]$. For each model, the hidden sizes and final-layer embedding sizes are 128. AT has 4 decoder layers to match the size of \textsc{Bert}.

\textbf{Results.} Figure \ref{fig:dist} visualizes how well the AT classifier and LDA predict topics as mixtures. It shows the recovered topic distributions on several validation datapoints when $\alpha=0.5$, and suggests that our method is successful at capturing both top-topic accuracy and the topic distribution spread.

Quantitative performance results are summarized in Table \ref{tab:toy-lm-lda}. In general, all four models demonstrate success at recovering latent topics by returning high accuracy and low loss on at least the easiest setting (i.e., $\alpha=0.5$). Between AT and \textsc{Bert}, we find that the probe on AT is able to infer latent topic structures in more difficult tasks (i.e., $\alpha=0.8, 1$) whereas the probe on \textsc{Bert} shows deteriorating performance. LDA outperforms both AT and \textsc{Bert} as expected because it is specified exactly to learn a dataset generated by the other manually initialized LDA. However, the strong WE performance suggests that the topic probe is able to predict topics mainly from stand-alone words. This raises the question whether AT and \textsc{Bert} are learning an underlying statistical model, or are simply uniquely embedding each word and making the probe mainly responsible for topic recovery.

\begin{table*}[h!]
    \centering
    \caption{20NG topic prediction performance based on different LLMs. \textit{Trained LLMs substantially outperform the null GPT-2 model, supporting the hypothesis that the training process encourages LLMs to implicitly develop topic models. Autoregressive models,  separated by line, statistically significantly outperform non-autoregressive models.}}
    \resizebox{1.9\columnwidth}{!}{
    \begin{tabular}{cccccccc}
    \toprule 
    & & & $K=20$ & & & $K=100$ & \\
    Model & Parameters & Accuracy $\uparrow$ & L2 loss $\downarrow$ & Tot. var. loss $\downarrow$ & Accuracy $\uparrow$ & L2 loss $\downarrow$ & Tot. var. loss $\downarrow$ \\
    \midrule
       \textsc{Gpt-2}  & 124M & $61.4\% \pm 1.5\%$ & $0.106 \pm 0.002$ & $0.211 \pm 0.001$ & $42.3\% \pm 2.4\%$ & $0.097 \pm 0$ & $0.192 \pm 0.001$\\
       \textsc{Gpt-2-medium} & 355M & $62.6\% \pm 1.7\%$ & $0.104 \pm 0.002$ & $0.209 \pm 0.002$ & $42.9\% \pm 2.4\%$ & $0.096 \pm 0$ & $0.19 \pm 0.001$\\
       \textsc{Gpt-2-large} & 774M & $62.4\% \pm 1.8\%$ & $0.102 \pm 0.002$ & $0.208 \pm 0.002$ & $43.1\% \pm 2.3\%$ & $0.095 \pm 0.001$ & $0.189 \pm 0$\\
       \textsc{Llama 2} & 7B & $62.6\% \pm 1.7\%$ & $0.101 \pm 0.002$ & $0.206 \pm 0.002$ & $43.3\% \pm 2.4\%$ & $0.095 \pm 0.001$ & $0.189 \pm 0.001$\\
       \textsc{Llama 2-chat} & 7B & $62.9\% \pm 1.7\%$ & $0.102 \pm 0.002$ & $0.207 \pm 0.002$ & $43.2\% \pm 2.5\%$ & $0.095 \pm 0.001$ & $0.189 \pm 0$\\
    \midrule
       \textsc{Bert} & 110M & $56.3\% \pm 1.5\%$ & $0.113 \pm 0.003$ & $0.222 \pm 0.003$ & $38.6\% \pm 2.5\%$ & $0.1 \pm 0.001$ & $0.191 \pm 0.001$\\
       \textsc{Bert-large} & 336M & $55.2\% \pm 1.2\%$ & $0.116 \pm 0.002$ & $0.226 \pm 0.003$ & $38.9\% \pm 2.9\%$ & $0.1 \pm 0.001$ & $0.191 \pm 0.001$\\
    \midrule
       Null \textsc{Gpt-2} & 124M & $27.3\% \pm 1\%$ & $0.209 \pm 0.003$ & $0.322 \pm 0.005$ & $13.8\% \pm 1.7\%$ & $0.145 \pm 0.001$ & $0.248 \pm 0.003$\\
    \bottomrule
    \end{tabular}}
    \label{tab:20ng}
\end{table*}

\begin{table*}[h!]
    \centering
    \caption{Wikitext-103 topic prediction performance based on different LLMs. \textit{On the accuracy metric, the \textsc{Llama2} series have ambiguous performance compared with MLMs, and aside from those, autoregressive models, separated by line, statistically significantly outperform non-autoregressive models.}}
    \resizebox{1.9\columnwidth}{!}{
    \begin{tabular}{cccccccc}
    \toprule 
    & & & $K=20$ & & & $K=100$ & \\
    Model & Parameters & Accuracy $\uparrow$ & L2 loss $\downarrow$ & Tot. var. loss $\downarrow$ & Accuracy $\uparrow$ & L2 loss $\downarrow$ & Tot. var. loss $\downarrow$ \\
    \midrule
       \textsc{Gpt-2}  & 124M & $86.7\% \pm 0.5\%$ & $0.025 \pm 0$ & $0.098 \pm 0$ & $73.9\% \pm 2.3\%$ & $0.026 \pm 0.001$ & $0.089 \pm 0.002$\\
       \textsc{Gpt-2-medium} & 355M & $88.2\% \pm 0.6\%$ & $0.024 \pm 0$ & $0.097 \pm 0.001$ & $74.2\% \pm 1.3\%$ & $0.025 \pm 0$ & $0.097 \pm 0.002$\\
       \textsc{Gpt-2-large} & 774M & $88.5\% \pm 0.8\%$ & $0.023 \pm 0$ & $0.094 \pm 0.001$ & $74.2\% \pm 1.4\%$ & $0.025 \pm 0$ & $0.088 \pm 0.001$\\
       \textsc{Llama 2} & 7B & $87.3\% \pm 1.7\%$ & $0.023 \pm 0$ & $0.091 \pm 0.001$ & $70.4\% \pm 1.1\%$ & $0.026 \pm 0$ & $0.09 \pm 0.001$\\
       \textsc{Llama 2-chat} & 7B & $85.3\% \pm 0.7\%$ & $0.024 \pm 0$ & $0.094 \pm 0$ & $69.9\% \pm 1\%$ & $0.026 \pm 0$ & $0.09 \pm 0$\\
    \midrule
       \textsc{Bert} & 110M & $84.9\% \pm 1.1\%$ & $0.027 \pm 0$ & $0.103 \pm 0.001$ & $72.4\% \pm 1.4\%$ & $0.029 \pm 0$ & $0.097 \pm 0.002$\\
       \textsc{Bert-large} & 336M & $85.4\% \pm 1.7\%$ & $0.03 \pm 0$ & $0.111 \pm 0$ & $72.1\% \pm 0.9\%$ & $0.031 \pm 0$ & $0.104 \pm 0$\\
    \midrule
       Null \textsc{Gpt-2} & 124M & $58.1\% \pm 1.8\%$ & $0.121 \pm 0.003$ & $0.247 \pm 0.006$ & $32.9\% \pm 3.2\%$ & $0.099 \pm 0.003$ & $0.195 \pm 0.007$\\
    \bottomrule
    \end{tabular}}
    \label{tab:wiki}
\end{table*}

\subsubsection{Controlling for Probe Performance}

In this section, we conduct control experiments that suggest that language models learn an underlying statistical model, making them---rather than the topic probes---mainly responsible for successful topic recovery. 
If AT or \textsc{Bert} performs well just because it gives each word a unique embedding from which a trained probe suffices to recover the topic mixture, then a probe on top of the language model should additionally predict topic mixtures from a different underlying topic model than the one that generated the language model's training data. To test whether this is the case, we generate five datasets using five distinct topic models under the setting of $\alpha=0.5$. 
One AT and one \textsc{Bert} models are trained on each dataset. On each model, five probes are used to predict topics from each of the five datasets. It is expected this will result in weak performance on datasets generated by unrelated topic models.
Results are shown in Figure \ref{fig:heatmap-synthetic}. AT shows strong distinction between predicting its own dataset versus predicting datasets from other topic models, whereas this distinction is present but weaker for \textsc{Bert}.


\subsection{Natural Corpora}

As predicted by our analysis based on de Finetti’s theorem, we have shown that latent topics can be decoded from LLMs trained on fully exchangeable texts. But it remains to be seen whether they still encode topics on natural texts where exchangeability does not hold. We presented several arguments for why this might be so. 
This section provides the necessary empirical tests by analyzing LLMs trained on natural language.

\begin{figure}[h]
    \centering
    \begin{subfigure}[b]{0.4\textwidth}
        \centering
        \includegraphics[width=\textwidth]{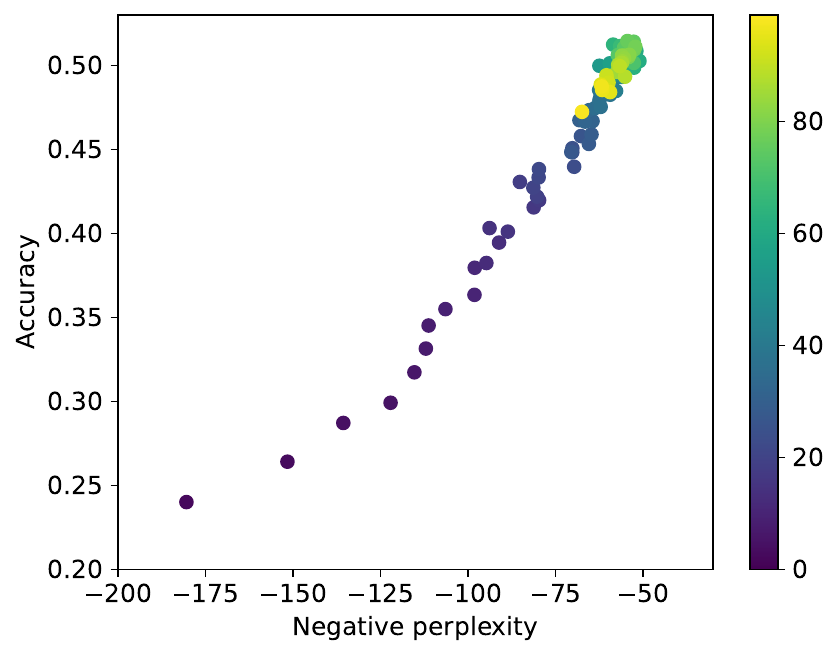}
        \caption{Accuracy vs. negative perplexity}
        \label{fig:token-acc-perp}
    \end{subfigure}
    \hfill
    \caption{20NG probe classification performance (accuracy) vs.~ negative perplexity measured at 100 different tokens. The dots are colored by the position percentile. {\it Probe performance increases with lower perplexity}.}
    \label{fig:token-perp}
\end{figure}

\textbf{Datasets}. We first use a natural corpus, \textit{20Newsgroups} (20NG), that can be naturally grouped into twenty topics. The corpus is a collection of eighteen thousand posts, and these posts are written in a style similar to informal emails. 

To contrast with the informal language style in 20NG, we also apply our analysis to WikiText-103 \citep{merity2016pointer}. This dataset consists of over 100 million tokens sourced from the selection of verified articles on Wikipedia that are classified as Good and Featured. 

\textbf{Setup.} We begin by training LDA models across three random seeds on each dataset, and then use pretrained large language models as our LLM. Specifically, the LLMs are \textsc{Gpt-2}, \textsc{Gpt-2-medium}, \textsc{Gpt-2-large}, \textsc{Llama 2}, \textsc{Llama 2-chat}, \textsc{Bert}, and \textsc{Bert-large}. The classifiers trained on top of these LLMs target LDA-learned topic mixtures with $K=20, 100$. For the models, we additionally include a randomly initialized \textsc{Gpt-2}, called Null \textsc{Gpt-2}, to differentiate model performance from probe performance.

\textbf{Implementation.} During topic probe training, learning rates are searched in $[10^{-5}, 10^{-3}]$, and L2 weight-decay in $[3.4\cdot10^{-5}, 3.4]$. The Adam optimizer was used for all training processes. Considering that the first token may lead to a useful embedding that stands for the $\langle \text{CLS} \rangle$ token in pretrained \textsc{Bert} series, we searched across the first, last, and average embeddings. The performance metrics are the same as those used in the synthetic datasets of Section \ref{sec:4.1}.

\textbf{Topic prediction results.} Quantitative performance is shown in Table \ref{tab:20ng} for 20NG and in Table \ref{tab:wiki} for Wikitext-103. To predict mixtures of twenty topics, random guessing would yield a $5\%$ accuracy for $K=20$ or $1\%$ accuracy for $K=100$. The best-performing LLM on each dataset demonstrates success at encoding topic distributions by achieving $62.9\% / 43.2\%$ accuracy on 20NG, and $88.5\% / 74.2\%$ on WikiText-103---scores that are substantially higher than those for the null GPT-2 model. 
Similarly to the pattern in synthetic datasets, autoregressive LLMs, namely the \textsc{Gpt-2} and \textsc{Llama 2}, outperform MLMs, namely \textsc{Bert}. We also explore the options of using other LLM layers, as opposed to only the last layer, as their document representation. Results suggest that topic mixtures are encoded equally well by later layers and earlier layers.

\textbf{Probe performance and LLM perplexity.} To study the relationship between word-prediction quality and topic awareness, we analyze topic prediction accuracy at 100 different positions on the token embeddings and their corresponding LLM perplexity (Figure \ref{fig:token-perp}).  Each position is defined based on the corresponding percentile of the total document (each document has a different length). Topic probes are trained by each taking embeddings at a position percentile as input. Their respective accuracies and loss metrics are measured, and LLM perplexity at these tokens are also computed. We expect that as perplexity on a token increases, probe performance based on embedding taken from that token would decrease. This hypothesis is supported by the linear trend in Figure \ref{fig:token-perp}. To validate the relationship between perplexity and probe performance, we additionally control for the effect from token position and overall document perplexity using a mixed effects linear model (\ref{sec:appendix_lmer}), confirming their statistical significance.

\section{Discussion}
\label{sec:discussion}

Our results indicate that LLM embeddings do contain information that can be used to decode the topic representation of documents, consistent with our hypothesis motivated by de Finetti's theorem. However, there are several directions for future work.

\textbf{Unsupervised approach for topic probing.} Our analyses used a particular topic model as the target for our trained classifiers, but that topic model is unlikely to match the internal topic analysis that LLMs actually arrive at. One potential direction for future work would be to develop an unsupervised approach for determining what topics an LLM assumes, so that we would not be dependent on a specific separate LDA model.

\textbf{Connecting to latent variables besides topics.} We used de Finetti's theorem to connect to one type of latent generating distribution, namely topics. This same analysis may also explain some of the other latent variables that have been mentioned in the related work section. 
In particular, given the central role of exchangeability in our analysis, this analysis would most naturally be extended to other latent variables that do not depend heavily on word order, such as the author of the document \citep{andreas-2022-language} or the author's sentiment \citep{radford2017learning}. 

\textbf{Using Bayesian probabilistic models to interpret neural networks.} Our analyses demonstrate the applicability of Bayesian models in deep learning, serving as a foundation for interpreting the mechanisms used by language models. A prospective path of exploration involves extending such analyses to other Bayesian models for both text and other modalities, in conjunction with deep learning models of these modalities.


\section{Conclusion}
\label{sec:conclusion}

We have shown that LLMs trained on next-word prediction are also learners of topic mixtures. To explain this phenomenon, we connected the LLM objective with de Finetti's theorem: under the assumption of exchangeability, autoregressive prediction is equivalent to Bayesian inference on latent generating distributions, and we have argued that, although language in its full complexity is not exchangeable, we can approximate this equivalence for aspects of language for which order of words is not crucial (such as topic structure). Our results show that LDA can be used to drive hypotheseses and empirical evaluations of what LLMs learn. We hope that our experiments will pave the way for future work that continues to bridge the gap between Bayesian probabilistic models and deep neural networks.


\subsubsection*{Acknowledgements}
This work was supported by ONR grant number N00014-23-1-2510. TRS is supported by an NDSEG Fellowship. RTM is supported by the NSF SPRF program under Grant No.\ 2204152.



\renewcommand{\bibsection}{\subsubsection*{References}}
\bibliography{ref}
\bibliographystyle{plainnat}

\newpage
\appendix
\onecolumn
\section{Appendix}
\label{sec:appendix}

\subsection{Proof}
\label{sec:appendix_proof}

We shall prove the equivalence between the MLM objective and Bayesian inference in Equation \ref{eq:mlm-bayes}. The statement is that, given an exchangeable process $x_1, x_2, ..., x_n$,
\begin{align}
    \mathcal{L}_{MLM}(x_{1:N}) &:= \sum_{n \in M} \log p(x_n | x_{i,i\in U}) \\
    &= \sum_{n \in M} \log \int p(x_n | \theta)p(\theta|x_{i,i\in U})d\theta, \label{eq:mlm_bayes_appendix}
\end{align}
where $M$ is the set of masked indices and $U$ is the set of unmasked indices, and by construction $M \cap U = \emptyset$.

\paragraph{Proof.} We first prove the autoregressive version of the equivalence, which \citet{korshunova2018bruno} proposes for each individual term in the summation but briefly mentions why it is equivalent,
\begin{align}
    \log p(x_{1:N}) &= \log p(x_1) + \sum_{n=1}^{N-1} \log p(x_{n+1}|x_{1:n}) \nonumber \\
    &= \log p(x_1) + \sum_{n=1}^{N-1} \log \int_\theta p(x_{n+1}|\theta)p(\theta|x_{1:n})d\theta.
\end{align}
To do so, we shall prove the equivalence in each term in the summation, that is, the statement that
\begin{align}
    p(x_{n+1}|x_{1:n}) = \int_\theta p(x_{n+1}|\theta)p(\theta|x_{1:n})d\theta \label{eq:gpt_bayes_appendix}
\end{align}
for each $n>1$. First, de Finetti's theorem states that, under the same exchangeability condition,
\begin{align}
    p(x_{1:n+1}) = \int_{\theta} p(\theta)\prod_{i=1}^{n+1}p(x_i|\theta) d\theta,
\label{eq:definetti-appendix}
\end{align}
and also that each $x_i$ is conditionally independent given $\theta$ for all $i$. To show the equivalence, we first divide each side of Equation \ref{eq:definetti-appendix} by $p(x_{1:n})$, assuming that $p(x_{1:n}) > 0$. The left hand side becomes,
\begin{align}
    \frac{p(x_{1:n+1})}{p(x_{1:n})} = p(x_{n+1}|x_{1:n}).
\end{align}
The right hand side becomes,
\begin{align}
\int_{\theta} p(\theta)\frac{\prod_{i=1}^{n+1}p(x_i|\theta)}{p(x_{1:n})} d\theta &= \int_{\theta} p(\theta)\frac{p(x_{n+1}|\theta) p(x_{1:n}|\theta)}{p(x_{1:n})} d\theta \label{eq:001}\\
&= \int_{\theta} p(x_{n+1}|\theta) \frac{p(\theta)p(x_{1:n}|\theta)}{p(x_{1:n})} d\theta \\
&= \int_{\theta} p(x_{n+1}|\theta) p(\theta|x_{1:n} )d\theta. \label{eq:002}
\end{align}
Line \ref{eq:001} uses conditional independence to combine the product on $i=1$ through $n$, and line \ref{eq:002} uses Bayes rule. Therefore, because of Equation \ref{eq:definetti-appendix}, we prove the statement of Equation \ref{eq:gpt_bayes_appendix}.

The proof for MLM in Equation \ref{eq:mlm_bayes_appendix} can be shown by a simple extension. We shall use $x_U$ as short hand for $x_{i,i\in U}$. Using de Finetti's theorem, 
\begin{align}
    p(x_{\{n\} \cup U}) = \int_{\theta} p(\theta)\prod_{i \in \{n\} \cup U}p(x_i|\theta) d\theta,
\label{eq:definetti-appendix-mlm}
\end{align}
Dividing each side of the above equation by $p(x_U)$, the left hand side becomes,
\begin{align}
    \frac{p(x_{\{n\} \cup U})}{p(x_U)} = p(x_n|x_U).
\end{align}
The right hand side becomes,
\begin{align}
\int_{\theta} p(\theta)\frac{\prod_{i \in \{n\} \cup U}p(x_i|\theta)}{p(x_U)} d\theta &= \int_{\theta} p(\theta)\frac{p(x_{n}|\theta) p(x_U|\theta)}{p(x_U)} d\theta \label{eq:003}\\
&= \int_{\theta} p(x_{n}|\theta) \frac{p(\theta)p(x_{U}|\theta)}{p(x_{U})} d\theta \\
&= \int_{\theta} p(x_{n}|\theta) p(\theta|x_{U} )d\theta. \label{eq:004}
\end{align}
Same as in the proof for the autoregressive version, line \ref{eq:003} uses conditional independence to combine the product on $U$, and line \ref{eq:004} uses Bayes rule. Therefore, we have that $p(x_n | x_{i,i\in U}) = \int p(x_n | \theta)p(\theta|x_{i,i\in U})d\theta$.
Thus, we have shown the equivalence for each term in the summation of the original statement Equation \ref{eq:mlm_bayes_appendix}.

\subsection{Linear Mixed-Effects Model}
\label{sec:appendix_lmer}

We want to validate our hypothesis that the LLM's latent topic representation helps it predict individual tokens. While Fig.~\ref{fig:token-acc-perp} is suggestive of this relationship, here we use statistical testing to confirm it.

Concretely, we use a linear mixed-effects model to predict the per-token perplexity. We analyze 701,243 individual tokens from 20NG test corpus using \textsc{Gpt-2}. Perplexity naturally decreases as the LLM processes the document, so we include a fixed effect of token position and a random effect for the document itself; finally, we include the topic decoding accuracy (a binary 0 / 1 outcome based on the topic probe) as the variable of interest. We extract 100 tokens per document, stratified so they are evenly spaced, and represent the token position as the percent into the document,

\begin{equation}
    \text{perplexity} \sim \text{token\_position} + \text{topic\_accuracy} + (1 | \text{document\_id}).
\end{equation}

We find significant effects for both token position and topic accuracy,

\begin{table}[ht]
\centering
\begin{tabular}{lllrrrrr}
  \hline
Effect & Group & Term & Estimate & Std. Error & Statistic & DF & p-value \\ 
  \hline
fixed &  & (Intercept) & 4.65 & 0.01 & 413.28 & 21078.63 & <2e-16 \\ 
fixed &  & topic\_accuracy & -0.15 & 0.01 & -16.51 & 355354.64 & <2e-16 \\ 
fixed &  & token\_position & -0.78 & 0.01 & -58.47 & 696289.69 & <2e-16 \\ 
ran\_pars & document\_id & sd\_\_(Intercept) & 0.63 &  &  &  &  \\ 
ran\_pars & Residual & sd\_\_Observation & 3.22 &  &  &  &  \\ 
   \hline
\end{tabular}
\end{table}

Finally, we obtain a Variance Inflation Factor of 1.014742 between accuracy and token position, suggesting an acceptable degree of colinearity between the two variables.

\subsection{Implementational Details}
\label{sec:appendix_implementation}

Here we detail the hyperparameter setup for experiments. All computations for synthetic datasets are run on single Tesla T4 GPUs, and those for natural corpora are run on single A100 GPUs. 

\subsubsection{Synthetic Data}

Autoregressive transformer (AT) and \textsc{Bert} hyperparameters for training are given in Table \ref{tab:syn-appendix-1}. Dropout ratio for each model is set to $0.1$ (default). Configurations of \textsc{Bert} are identical to those of \textsc{Bert-tiny} from \citet{Turc2019WellReadSL}. AT hidden sizes and final layer embedding sizes are 128, same as \textsc{Bert-tiny}, and it uses four layers, resulting in 655,336 parameters. \text{Bert} has 608,747 parameters.

We train language models on 10,000 documents. Probes trained on language model embeddings of 1,000 documents that are unseen by language models, and probes are evaluated on additional 1,000 documents that are unseen by both language models and probes.

\begin{table}[]
    \centering
    \caption{Autoregressive transformer (AT) and \textsc{Bert} hyperparameters for training on the synthetic datasets.}
    \begin{tabular}{c c c}
    \toprule
        Parameter & Tuning range & Chosen value\\
    \midrule
        Batch-size & $[8,128]$ & $16$\\
        Learning rate & $[3\cdot 10^{-5}, 10^{-3}]$ & $10^{-4}$\\
    \bottomrule
    \end{tabular}
    \label{tab:syn-appendix-1}
\end{table}

Hyperparameters for probes on AT and \textsc{Bert} are given in Table \ref{tab:syn-appendix-2}.

\begin{table}[]
    \centering
    \caption{Probe hyperparameters for training on top of synthetic dataset language models.}
    \begin{tabular}{c c c}
    \toprule
        Parameter & Tuning range & Chosen value\\
    \midrule
        Batch-size & $[8,64]$ & $16$\\
        Learning rate & $[10^{-4}, 0.03]$ & $10^{-3}$\\
        Weight-decay & $[0, 3.4 \cdot 10^{-4}]$ & 0\\
        Embedding choice & $\{\text{First, Last, Average}\}$ & Last for AT / Average for \textsc{Bert}\\
    \bottomrule
    \end{tabular}
    \label{tab:syn-appendix-2}
\end{table}

\subsubsection{Natural Corpora}

Hyperparameters for probes on the LLMs are given in Table \ref{tab:nat-appendix-1} and Table \ref{tab:nat-appendix-2}. For 20NG, probe training and validation are run on 11,314 and 7,532 documents, respectively. For WikiText-103, probe training and validation are run on 28,475 and 60 documents, respectively. Both splits are derived directly from train-validation split provided by the dataset sources.

\begin{table}[h!]
    \centering
    \caption{Probe hyperparameters for training on top of \textsc{Gpt-2}, \textsc{Gpt-2-medium}, \textsc{Gpt-2-large}, \textsc{Bert}, and \textsc{Bert-large}.}
    \begin{tabular}{c c c}
    \toprule
        Parameter & Tuning range & Chosen value\\
    \midrule
        Batch-size & $\{128\}$ & $128$\\
        Learning rate & $[10^{-5}, 10^{-3}]$ & $3 \cdot 10^{-4}$\\
        Weight-decay & $[0, 3.4]$ & $3.4 \cdot 10^{-3}$\\
        Embedding choice & $\{\text{First, Last, Average}\}$ & Average\\
    \bottomrule
    \end{tabular}
    \label{tab:nat-appendix-1}
\end{table}

\begin{table}[h!]
    \centering
    \caption{Probe hyperparameters for training on top of \textsc{Llama 2} and \textsc{Llama 2-chat}.}
    \begin{tabular}{c c c}
    \toprule
        Parameter & Tuning range & Chosen value\\
    \midrule
        Batch-size & $\{128\}$ & $128$\\
        Learning rate & $[10^{-5}, 10^{-3}]$ & $10^{-4}$\\
        Weight-decay & $[0, 3.4]$ & 0.34\\
        Embedding choice & $\{\text{First, Last, Average}\}$ & Average\\
    \bottomrule
    \end{tabular}
    \label{tab:nat-appendix-2}
\end{table}

\end{document}